\documentclass[conference]{IEEEtran}
\usepackage{times}
\usepackage{multicol}
\usepackage{xcolor}
\usepackage[hyphens]{url}
\usepackage[bookmarks=true]{hyperref}
\usepackage[hyphenbreaks]{breakurl}

\usepackage[numbers,sort&compress]{natbib}
\usepackage{graphicx}
\usepackage{float}
\usepackage{dblfloatfix}
\usepackage{caption}

\usepackage{amsmath}
\usepackage{amssymb}
\usepackage{mathtools}
\usepackage{xfrac}
\usepackage{bm}

\usepackage{algpseudocode}
\usepackage{algorithm}

\algnewcommand\And{\textbf{and}}

\newcommand{\txtapprox}{\raisebox{0.5ex}{\texttildelow}}
\newcommand{\m}[1]{\begin{bmatrix}#1\end{bmatrix}} 
\newcommand{\mcomp}{\thinmuskip=3mu\medmuskip=3mu\thickmuskip=3mu}

\ifx10
	\newcommand{\urlinfo}{https://XXXXXXX.XX/home?projects_momav}
	\newcommand{\urlgit}{https://github.com/XXXXXX/momav}
	\newcommand{\authortext}{Author Names Omitted for Anonymous Review. Paper-ID: 30}
	\newcommand{\marco}{M\#\#\#\#\#}
\else
	\newcommand{\urlinfo}{https://marcoruggia.ch/home?projects_momav}
	\newcommand{\urlgit}{https://github.com/mruggia/momav}
	\newcommand{\authortext}{Marco Ruggia - Institute of Photonics and Robotics, University of Applied Sciences of the Grison, Switzerland}
	\newcommand{\marco}{Marco}
\fi
\begin{document}

\title{MOMAV: A highly symmetrical fully-actuated multirotor drone using optimizing control allocation}
\author{\authortext{}}
\maketitle

%%%%%%%%%%%%%%%%%%%%%%%%%%%
\begin{abstract}
	MOMAV (\marco{}'s Omnidirectional Micro Aerial Vehicle) is a multirotor drone that is fully actuated, meaning it can control its orientation independently of its position. MOMAV is also highly symmetrical, making its flight efficiency largely unaffected by its current orientation. These characteristics are achieved by a novel drone design where six rotor arms align with the vertices of an octahedron, and where each arm can actively rotate along its long axis. Various standout features of MOMAV are presented: The high flight efficiency compared to arm configuration of other fully-actuated drones, the design of an original rotating arm assembly featuring slip-rings used to enable continuous arm rotation, and a novel control allocation algorithm based on sequential quadratic programming (SQP) used to calculate throttle and arm-angle setpoints in flight. Flight tests have shown that MOMAV is able to achieve remarkably low mean position/orientation errors of 6.6mm, 2.1° ($\bm\sigma$: 3.0mm, 1.0°) when sweeping position setpoints, and 11.8mm, 3.3° ($\bm\sigma$: 8.6mm, 2.0°) when sweeping orientation setpoints.
\end{abstract}

%%%%%%%%%%%%%%%%%%%%%%%%%%%
\section{Introduction}
Fully-actuated aerial vehicles are able to exert forces and torques independently of each other, enabling them to decouple translation and rotation movements. Unlike an ordinary multicopter, they can for example accelerate forward without having to lean forward, or lean forward without having to accelerate. This characteristic has been found to be particularly useful in manipulation tasks where aerial vehicles need to be in contact with the environment \cite{2021_ollero_manipulators, 2020_rashad_fullyactuated_review}. Such drones could possibly also become useful as an alternative to gimbals for sensors in visual inspections and surveying applications.
\par
Most fully-actuated aerial vehicles are multicopters that fall in one of two categories. Fixed-tilt designs, where the rotor disks are fixedly tilted away from the horizontal plane, as in \cite{2023_howard_lynchpin, 2016_brescianini_octcu, 2018_park_odar, 2020_hamadi_o7plus, 2022_flores_tiltedhex, 2021_ma_ams, 2019_rashad_hamiltonian_control, 2018_tognon_heptw, 2017_florentin_quad4hor, 2017_lei_cohexcd, 2013_jiang_hexc, 2012_toratani_hexdtet, 2015_nikou_heptf}, and variable-tilt designs where the rotor disks can be actively tilted, as in \cite{2019_bodie_omav, 2018_kamel_voliro, 2023_cuniatro_voliro2, 2021_buzzatto_unicopter, 2020_xu_bicopter, 2020_lee_t3multirotor, 2020_zheng_tiltdrone, 2016_ryll_fasthex, 2016_odelga_quadvcdc, 2015_kastelan_tricopter, 2015_moutinho_tiltquadrotor, 2015_ryll_quadvc, 2013_sequigasco_quadvcd, 2018_junaid_dualtiltquad}. In both cases at least six actuators are needed (either propeller motors or tilt servos) to control the six degrees of freedom of the drone (three for position and three for orientation).
\par
\begin{figure}[t!]
	\includegraphics[width=\columnwidth]{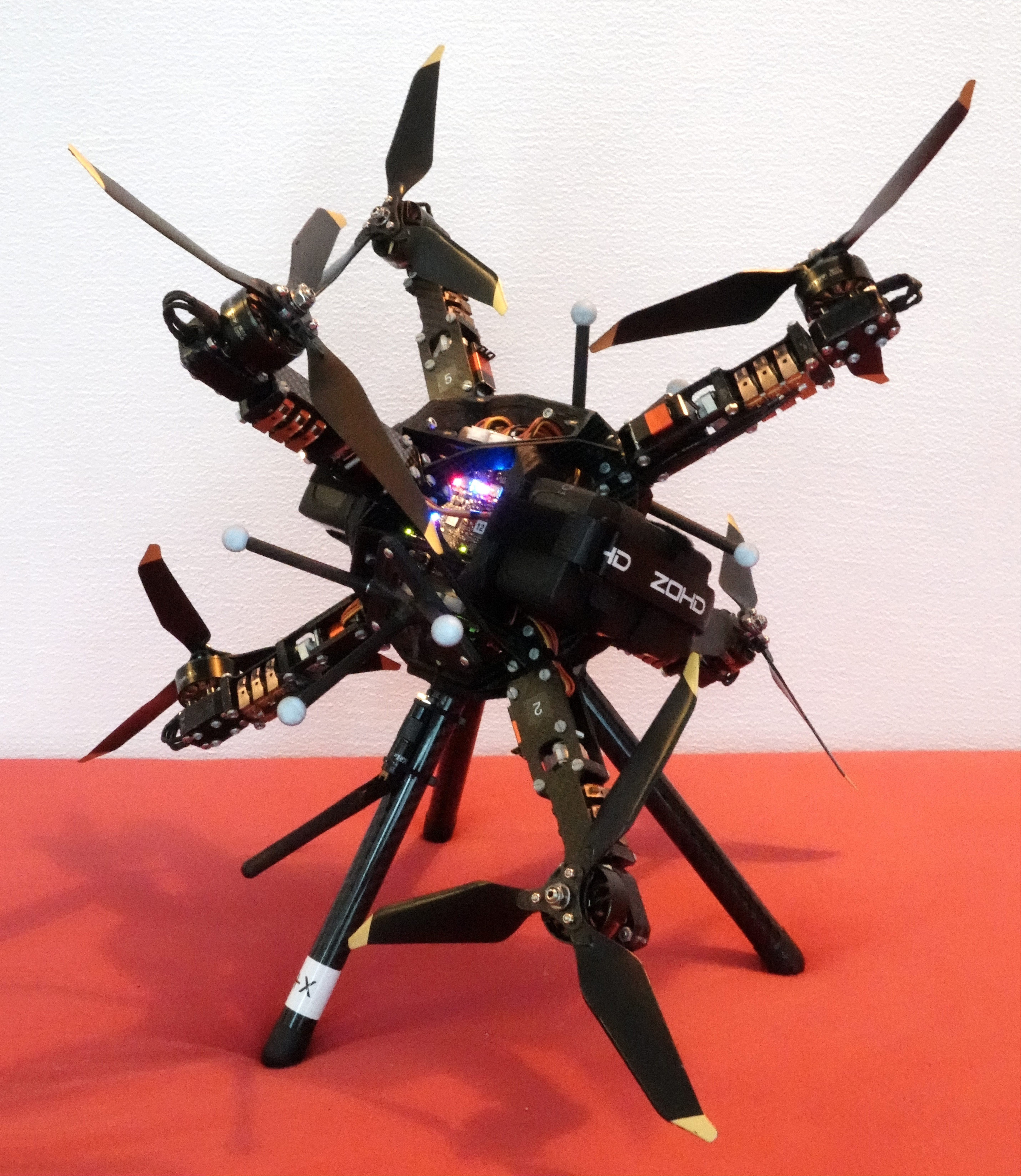}
	\caption{Prototype drone \textit{MOMAV}}
	\label{fig:0_drone}
	\vspace{-6pt}
\end{figure}

The here presented variable-tilt drone named MOMAV \mbox{(Fig. \ref{fig:0_drone})} features six propellers and six arm-angle servos, totaling twelve actuators. This exceeds the minimum of six, making the drone not only fully-actuated, but over-actuated. Thus, the spare actuators can be used to optimize additional objectives. Concretely, the additional objective of MOMAV is to increase flight efficiency independently of its current orientation. This is achieved through the choice of a novel highly symmetric arm configuration, and with help of a novel sequential quadratic programming (SQP) based control allocation algorithm.
\par
Additionally, this drone introduces an original rotating arm design, that connects the propeller motors in a way that enables continuous arm rotation using slip-rings. It serves to simplify the constraints placed on the control allocation. Not doing so would require a more sophisticated solution to prevent the cables powering the propeller motors from winding up around the arms after multiple rotations.

%%%%%%%%%%%%%%%%%%%%%%%%%%%
\section{Arm Configuration}
\begin{figure*}[t!]
	\includegraphics[width=\textwidth]{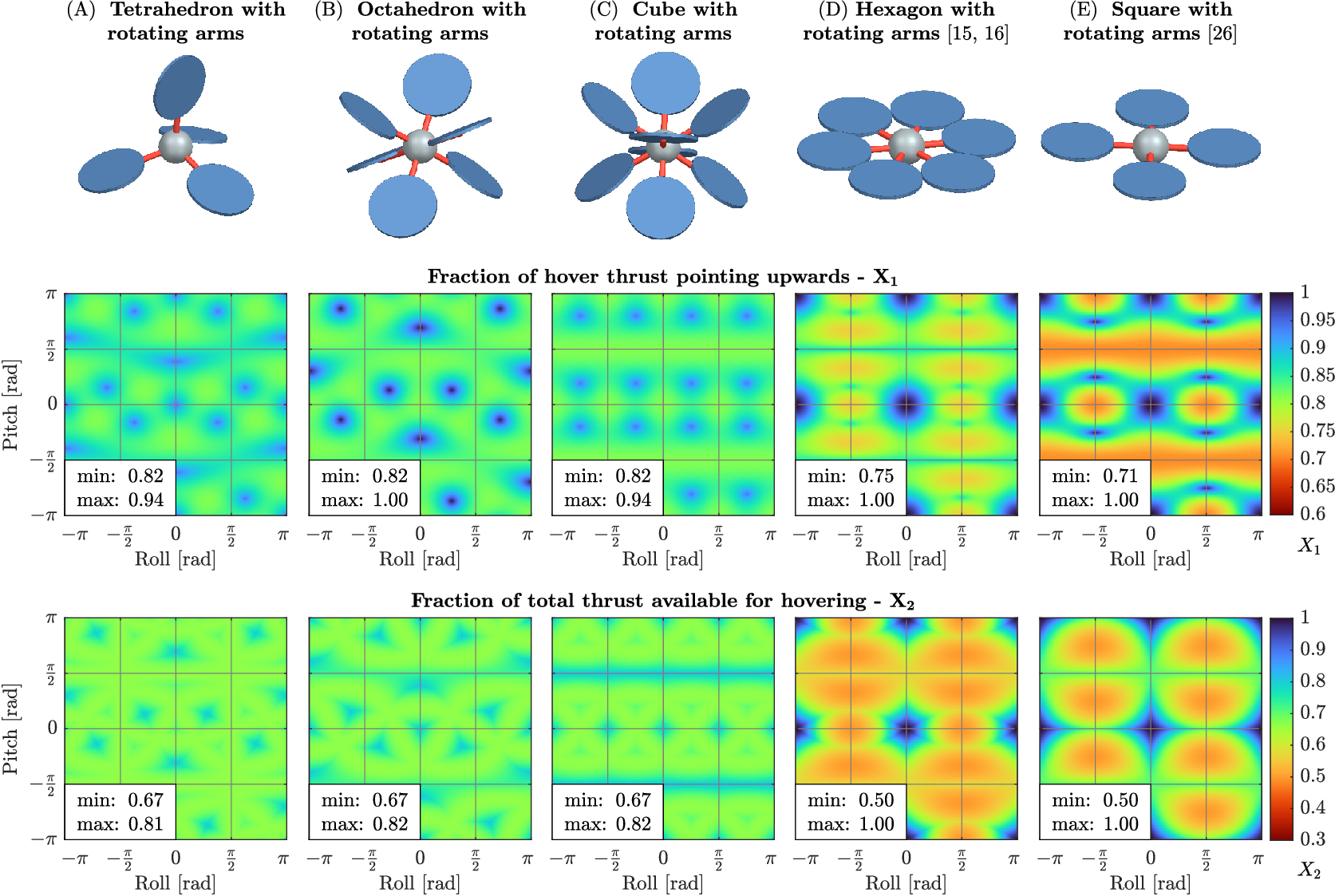}
	\centering
	\caption{Flight efficiency $X_1$, $X_2$ depending on orientation for a selection of drones with rotating arms}
	\label{fig:1_geometry_analysis1}
	\vspace{-4pt}
\end{figure*}
Many arm configurations for fully-actuated drones featuring ingenious mechanical solutions have been proposed. The vast majority of them \cite{2022_flores_tiltedhex, 2021_ma_ams, 2019_rashad_hamiltonian_control, 2018_tognon_heptw, 2017_florentin_quad4hor, 2017_lei_cohexcd, 2013_jiang_hexc, 2023_cuniatro_voliro2, 2021_buzzatto_unicopter, 2020_xu_bicopter, 2020_lee_t3multirotor, 2020_zheng_tiltdrone, 2016_ryll_fasthex, 2016_odelga_quadvcdc} have strict limitations on the orientations they can achieve. Drones capable of operating across the entire range of orientations have been shown to work only for two types of designs: Variable-tilt planar multicopter configurations where each arm can actively rotate along its long axis \cite{2019_bodie_omav, 2018_kamel_voliro}, and fixed-tilt multicopters with three-dimensionally positioned propellers \cite{2023_howard_lynchpin, 2016_brescianini_octcu, 2018_park_odar, 2020_hamadi_o7plus}. This project combines the concepts by implementing a hexacopter with arms aligned to the vertices of an octahedron (3D), where each arm can rotate. Such a configuration will be shown to have a consistently high efficiency across all orientations.
\par
Two metrics are calculated to compare flight efficiencies of various arm configurations. $X_1$ is the fraction of thrust directed upwards during hovering flight (Eq. \ref{eq:1_config_x1}). The remainder ${\mcomp (1-X_1)}$ is thrust, that is wasted by motors having to push against each other instead of upwards. $X_2$ on the other hand is the fraction of total thrust (sum of all maximal motor thrusts) that is available for hovering (Eq. \ref{eq:1_config_x2}). For example, a drone at $\mcomp X_2=0.5$ needs to be equipped with enough total thrust to lift at least twice its weight in order to hover. Both $X_1$ and $X_2$ are calculated by minimizing the sum of squared thrusts $f_i$ for arm configurations that exceed the six degrees of freedom needed to impose the hover condition (Eq. \ref{eq:1_config_metric}).
\par
\vspace{-10pt}
\begin{align}
	\phantom{\left(f_1^*,...\,,f_N^*\right) =}
	&\begin{aligned}
		\mathllap{\left(f_1^*,...\,,f_N^*\right)} =& \ \mathrm{\mathbf{argmin}} \ \textstyle\sum_{i=1}^{N} \|f_i\|^2 \ \ \mathrm{\mathbf{subject\ to:}}  \\
		&\ \textrm{\textit{constraints on}} \ f_i/\|f_i\| \ \textrm{\textit{directions}} \\
		&\ \textstyle\sum_{i=1}^{N} f_i = m \left[0,0,g\right]^T \\
		&\ \textstyle\sum_{i=1}^{N} r_i \times f_i = I \left[0,0,0\right]^T
		\label{eq:1_config_metric}
	\end{aligned}\\[10pt]
	&\begin{aligned}
		\mathllap{X_1} =&\ \frac{ \textstyle\sum_{i=1}^{N} f_i^* \cdot \left[0,0,1\right]^T }{ \textstyle\sum_{i=1}^{N} \|f_i^*\| } 
		\label{eq:1_config_x1}
	\end{aligned}\\[10pt]
	&\begin{aligned}
		\mathllap{X_2} =&\ \frac{ mg }{ \mathrm{max}\!\left(\|f_1^*\|,...\,,\|f_N^*\|\right) N } 
		\label{eq:1_config_x2}
	\end{aligned}
\end{align}
\par
By substituting a unitless thrust $f_i'=f_i/(mg)$ it can be easily shown that the motor thrusts at hover $f_i^*$ depend linearly on the drone mass $m$ and the gravitational acceleration $g$, which results in $X_1$ and $X_2$ being independent of both $m$ and $g$. The moment of inertia $I$ also plays no role. Only the constraints on the thrust directions $f_i/\|f_i\|$ of each motor and the motor locations $r_i$ (up to a scale factor) affect $X_1$, $X_2$.
\par
\begin{figure*}[t!]
	\includegraphics[width=\textwidth]{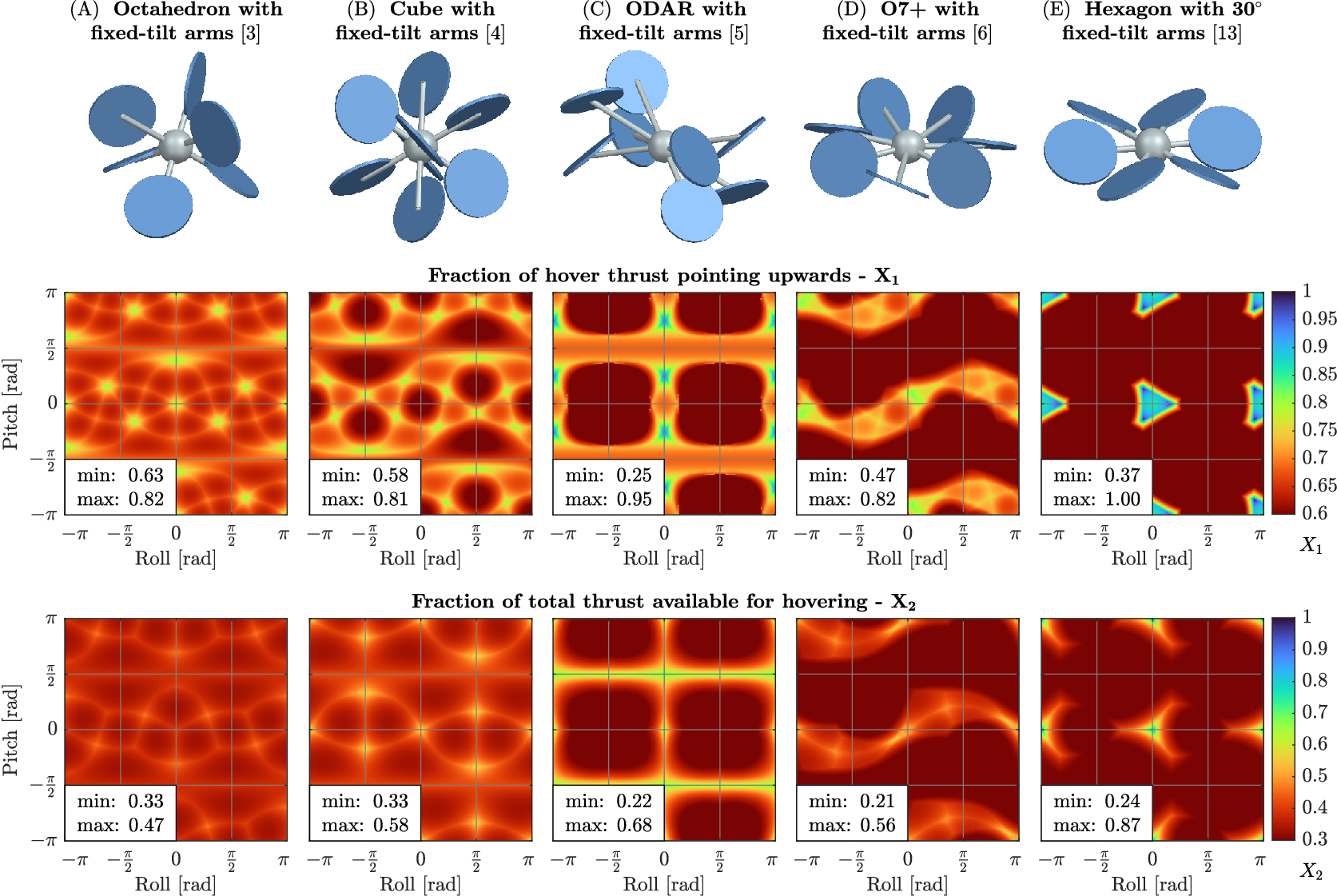}
	\centering
	\caption{Flight efficiency $X_1$, $X_2$ depending on orientation for a selection of drones with fixed-tilt arms}
	\label{fig:1_geometry_analysis2}
	\vspace{-4pt}
\end{figure*}
\par
Evaluating $X_1$ and $X_2$ across orientations for various configurations featuring rotating arms shows that the proposed octahedron with rotating arms (Fig. \hyperref[fig:1_geometry_analysis1]{\ref*{fig:1_geometry_analysis1}B}) achieves $\mcomp X_1\in[0.82,1.00]$, $\mcomp X_2\in[0.67,0.82]$, and that tetrahedra \mbox{(Fig. \hyperref[fig:1_geometry_analysis1]{\ref*{fig:1_geometry_analysis1}A})} and cubes (Fig. \hyperref[fig:1_geometry_analysis1]{\ref*{fig:1_geometry_analysis1}C}) with rotating arms perform very similarly to it. For the octahedron, $X_1$ is minimal when two opposing faces align with the vertical axis. In that case $\mcomp (1-X_{1, \mathrm{min}})=18\%$ of thrust is wasted to propellers having to push against each other. Conversely, $X_2$ is minimal when two opposing vertices align with the vertical axis. In that case only 4 of 6 motors are used for hovering, so at least $\mcomp (1-X_{2, \mathrm{min}})=2/6=33\%$ of total thrust must be kept as surplus in the design.
\par
Comparatively, a hexagon with rotating arms (Fig. \hyperref[fig:1_geometry_analysis1]{\ref*{fig:1_geometry_analysis1}D}, like in \cite{2019_bodie_omav, 2018_kamel_voliro}) reaches worse minima at $\mcomp X_1\in[0.75,1.00]$, $\mcomp X_2\in[0.50,1.00]$, and a square with rotating arms (Fig. \hyperref[fig:1_geometry_analysis1]{\ref*{fig:1_geometry_analysis1}E}, like in \cite{2015_ryll_quadvc}) is marginally worse still. Although, the latter was never shown to be capable of flying in arbitrary orientations. As expected, designs using fixed-tilt arms (Fig. \hyperref[fig:1_geometry_analysis2]{\ref*{fig:1_geometry_analysis2}A}--\hyperref[fig:1_geometry_analysis2]{\ref*{fig:1_geometry_analysis2}D}, like  in \cite{2023_howard_lynchpin, 2016_brescianini_octcu, 2018_park_odar, 2020_hamadi_o7plus}) show significantly worse $X_1$ and $X_2$ efficiencies across all orientations. Notably, the ODAR configuration \mbox{(Fig. \hyperref[fig:1_geometry_analysis1]{\ref*{fig:1_geometry_analysis1}C}}, like in \cite{2018_park_odar}), which was optimized for wrench generation across orientations, performs poorly with regards to efficiency. A hexagon with alternately 30° tilted arms (Fig. \hyperref[fig:1_geometry_analysis2]{\ref*{fig:1_geometry_analysis2}E} like in \cite{2013_jiang_hexc}) is also included in the analysis since it is a common configuration for fully-actuated drones, although no such design is currently capable of flying in arbitrary orientations.
\par
These results clearly indicate a theoretical efficiency advantage across orientations for fully-actuated drones using rotating arms in a highly symmetrical 3D configuration. These configurations include tetrahedra, octahedra and cubes, none of which have ever been successfully implemented before. Since all of them feature roughly the same  $X_1$ and $X_2$ efficiency, the octahedron is selected for further development as it is found to be the easier shape to manufacture.
\par
\begin{figure}
	\includegraphics[width=\columnwidth]{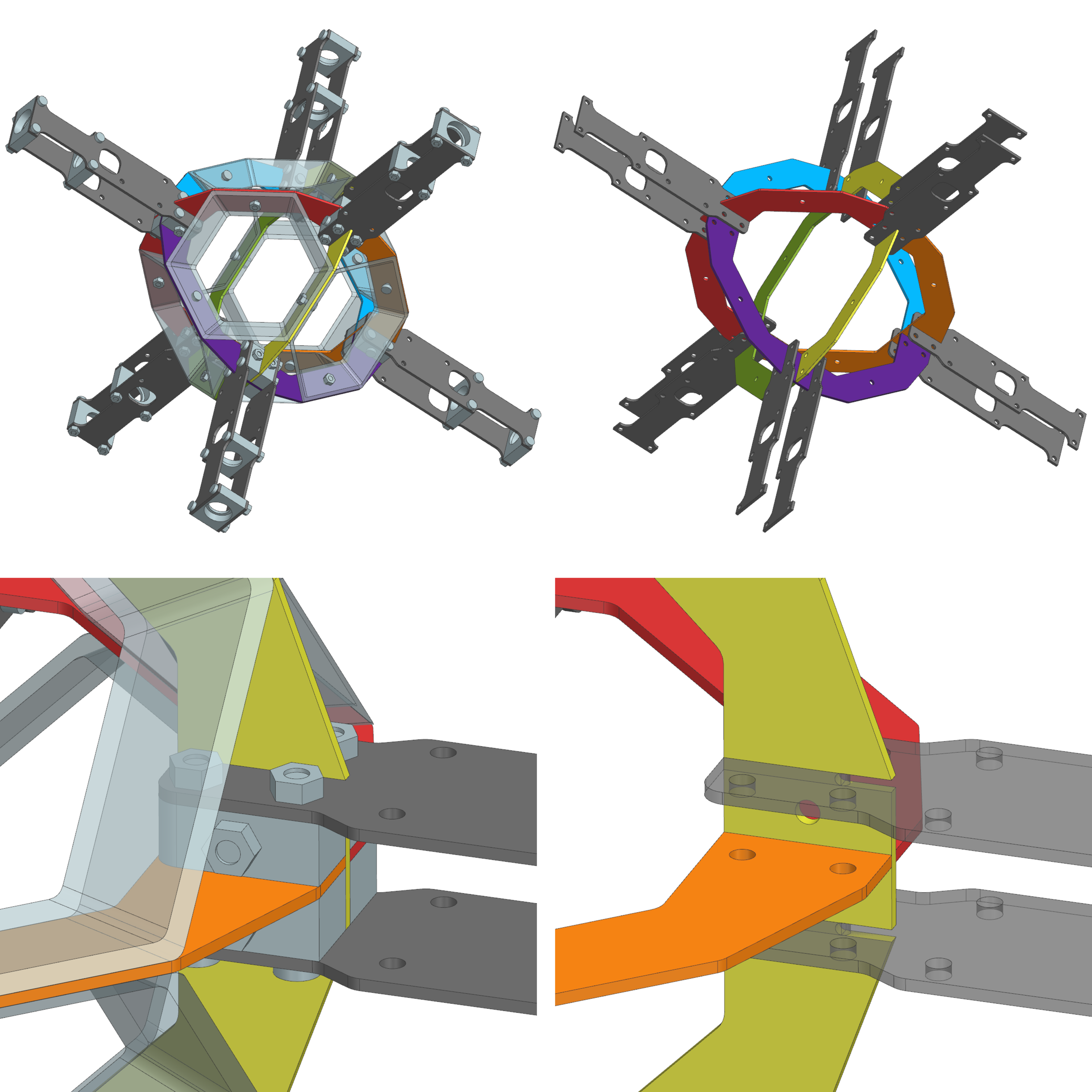}
	\centering
	\caption{Views of the octahedral body implementation showing details of the joints that connect the arms to the body}
	\label{fig:1_geometry_cad}
	\vspace{6pt}
\end{figure}

In practice, the octahedral body of MOMAV (Fig. \ref{fig:1_geometry_cad}) is constructed using carbon fiber sheets interlocking with halved joints. Additional rigidity is gained by connecting the edges of each face with hexagons, creating a truncated octahedron shape. Relying mainly on interlocking joints for connections results in a very rigid construction with no noticeable play.
\par

%%%%%%%%%%%%%%%%%%%%%%%%%%%
\section{Rotating Arm Assembly}
\begin{figure}[t!]
	\includegraphics[width=\columnwidth]{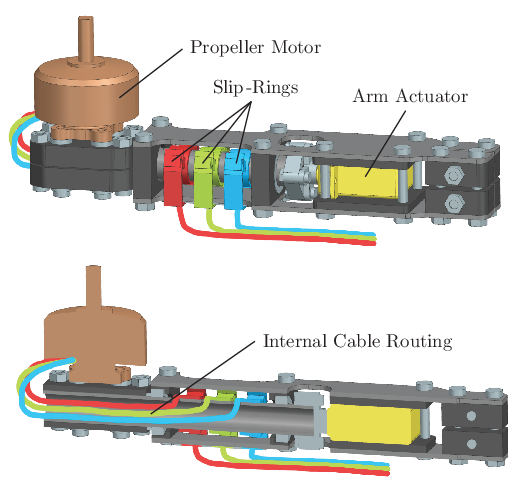}
	\centering
	\vspace{-14pt}
	\caption{Rotating arm assembly with a section view showing internal wiring of the three motor phases}
	\label{fig:2_arm_assembly}
	\vspace{-2pt}
\end{figure}

A second novel contribution of MOMAV are its arm assemblies (Fig. \ref{fig:2_arm_assembly}). Their function is to enable control of the arm-angles allowing for an arbitrary number of revolutions, and to transmit power to the propeller motors while doing so. These two functions can be challenging to implement. Here, solutions to both are proposed, analyzed, and then implemented on the MOMAV drone.
\begin{figure}[t!]
	\includegraphics[width=\columnwidth]{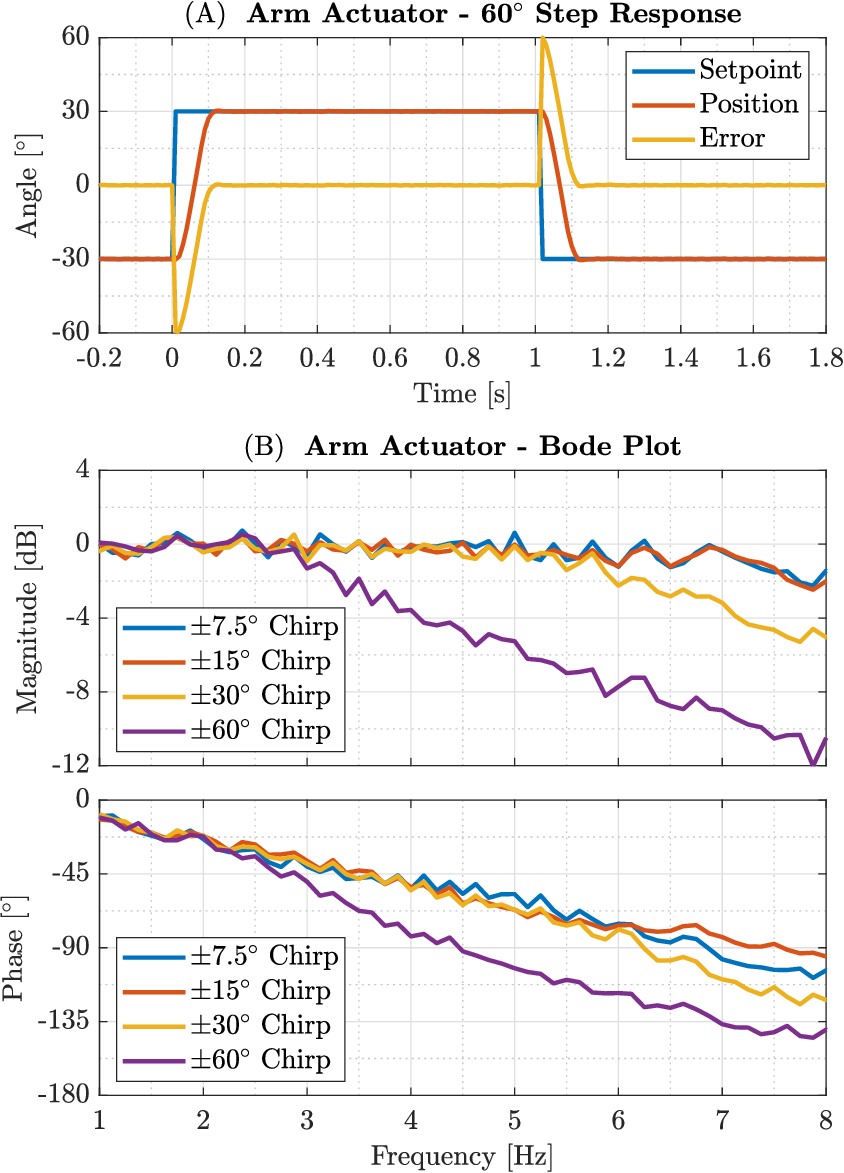}
	\centering
	\caption{Arm actuator step responses (A) and frequency response (B) with propeller motor running at \txtapprox12\,N thrust}
	\label{fig:2_servo_dynamics}
	\vspace{-4pt}
\end{figure}
\begin{figure*}[t!]
	\includegraphics[width=\textwidth]{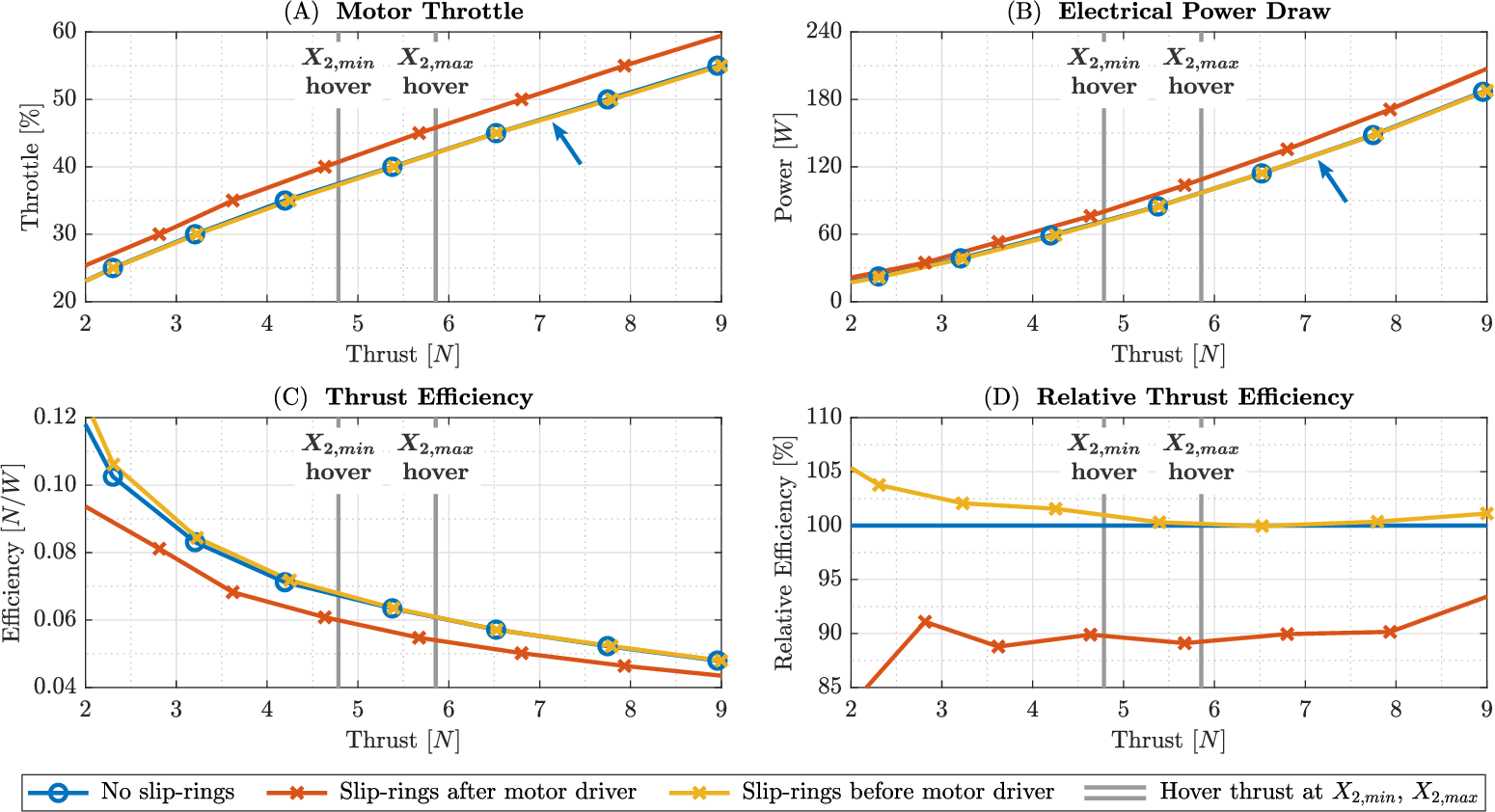}
	\centering
	\vspace{-4pt}
	\caption{Slip-ring efficiency analysis performed on a single arm in a thrust measuring stand for two different slip-ring placements}
	\label{fig:2_slipring_loss}
	\vspace{-4pt}
\end{figure*}

%%%%%%%%%%%%%%%%%%%%%%%%%%%
\subsection{Arm Actuator}
While some commercial solutions exist that are potentially suitable as arm actuators, none are found that fulfill all requirements on size, torque, velocity, accuracy, and rotation range demanded by this project. Instead, an existing servo motor, the \textit{KST MS325}, is modified to fit the requirements. 
\par
The only shortcoming of the \textit{KST MS325} servo is its 200° rotation angle limit. Otherwise, it features an outstanding 0.52\,Nm stall torque and 2.4\,rev/sec no-load speed for its 23$\times$12$\times$28\,mm size. This servo motor is also particular in that it features an \textit{AMS AS5600} magnetic encoder to measure angular position instead of the much more prevalent potentiometers. The magnetic encoder makes continuous rotation mechanically possible and only prevented by the built-in driver. So, a modification is performed by swapping the built-in driver for an \textit{Atmel SAM D21} microcontroller and a \textit{TI DRV8870} H-bridge. A PID control loop featuring \textit{proportional on measurement} (PoM) \cite{2017_beauregard_pom} is implemented on the microcontroller with the goal of avoiding the overshoot typically present on integrating systems like the DC motor used inside the servo.
\par
To evaluate the arm actuator a series of tests is performed while the propeller motor is running and producing approximately 12\,N of thrust. A step response test (Fig. \hyperref[fig:2_servo_dynamics]{\ref*{fig:2_servo_dynamics}A}) shows that the arm actuator is able to reach a setpoint angle precisely and without overshoot, being mostly only limited by a \txtapprox2.4\,rev/sec velocity saturation. 
\par
Under the same conditions, chirp signals of various amplitudes are tested, and frequency domain analysis performed by means of bode plots (Fig. \hyperref[fig:2_servo_dynamics]{\ref*{fig:2_servo_dynamics}B}). These show a magnitude fall-off with increasing chirp amplitude consistent with the previously noted velocity saturation, and a linear phase fall-off consistent with a time-delay of \txtapprox36\,ms.
\par

%%%%%%%%%%%%%%%%%%%%%%%%%%%
\subsection{Slip-Rings}
The second challenge solved by the rotating arm assembly is to transmit power from the stationary part of the arm to the rotating propeller motor. Slip-rings are proposed as a novel way to prevent the motor cables from winding up around the arms over time. Like for the arm actuators, it is exceedingly difficult to buy a slip-ring of suitable size and current rating for this task. Instead a solution is proposed, that uses brushes harvested from the ubiquitous \textit{RS550} DC motors.
\par
A pair of two brushes is used for each of the three motor phases. Each pair pushes against a copper ring placed on the rotating part of the arm. Wires then connect from the inside of these rings to the propeller motor (Fig. \ref{fig:2_arm_assembly}). 
\par
The efficiency of the slip-rings is evaluated with a single arm in a thrust measuring stand. Two configurations are analyzed: One with the motor driver placed before the slip-rings, as described previously, and one with the motor driver placed after the slip-rings, such that it lies on the rotating part of the arm. In that configuration direct current from the battery is transmitted through only two of the three slip-rings.
\par
The same motor and propeller are used as in the final MOMAV drone: A \textit{T-Motor F100 2810 1100KV} motor and a custom 3-bladed \textit{DJIMavic Pro 8331} propeller. A \textit{Tekko32 F4 Metal ESC 65A} is used as the motor driver and a \textit{Bota Medusa Serial} force-torque-sensor is used to measure thrust. The propeller motor is powered by two \textit{RND 320-KWR103} laboratory power supplies in parallel (2$\cdot$15\,A at 23.1\,V).
\par
The procedure for all tests consists of requesting constant throttles from the motor driver ranging from 20\% to 80\% in 5\% increments. During each increment the arm actuator spins slowly, performing 2 revolutions over 32 seconds. In this time the motor driver collects data on the motor current and the force-torque-sensor collects data on the produced thrust.
\par
The requested throttle is plotted against the measured thrust (Fig. \hyperref[fig:2_slipring_loss]{\ref*{fig:2_slipring_loss}A}). All other measurements are plotted against thrust instead of throttle to make comparisons between the configurations easier. The electrical power draw (Fig. \hyperref[fig:2_slipring_loss]{\ref*{fig:2_slipring_loss}B}) is calculated by multiplying the current measured by the motor driver with the 23.1\,V regulated supply voltage. Then, the thrust efficiency (Fig. \hyperref[fig:2_slipring_loss]{\ref*{fig:2_slipring_loss}C}) is calculated as the ratio of produced thrust per consumed electrical power. Finally, the relative efficiency \mbox{(Fig. \hyperref[fig:2_slipring_loss]{\ref*{fig:2_slipring_loss}D})} of both slip-ring configurations is calculated by dividing the thrust efficiency without slip-rings by the thrust efficiency with slip-rings for equal thrusts.
\par
Two notable thrust setpoints are marked in the plots. One at $\mcomp X_{2,min}\cdot mg/6 = 4.8N$ and one at $\mcomp X_{2,max} \cdot mg/6 = 5.9N$. The first corresponds to the thrust required by all 6 motors to hover when a face of the octahedral body aligns with the vertical axis. The second corresponds to the thrust required by 4 out of 6 motors to hover when a vertex of the octahedral body aligns with the vertical axis.
\par
In the region of hover thrusts, a relative thrust efficiency of 90\% is observed for the slip-ring configuration implemented on MOMAV, meaning that about 10\% of power is wasted due to the presence of the slip-rings. In comparison, the alternative slip-ring placement between the battery and the motor driver yields a relative thrust efficiency of roughly 100\%.
\par
Two possible reasons are noted that might explain these results. The current after the motor driver is increased because of the voltage drop set by the throttle (conservation of power). And the cables between driver and motor are substantially shorter in the alternative slip-ring placement test (2\,cm) compared to the two other tests (25\,cm). The total cable length between power supply and motor is however equal in all tests.
\par
These results show that the placement of slip-rings on MOMAV was a major design oversight, and that the same slip-ring design could be implemented without incurring a penalty in efficiency by placing them between the battery and the motor drivers. Although, this insight should at least prove useful for future applications of the here presented high-current slip-ring design.
\par
As for thermal considerations, the current and temperature at the slip-rings are analyzed under the most extreme tested conditions at 80\% throttle at the end of each test run. The temperature is measured with a \textit{SEEK Thermal Compact Pro} between two slip-rings at the rotating shaft. For slip-rings placed after the motor driver, 60\,°C at 27.2\,A are measured (3-phase current), and for slip-rings placed before the motor driver, 50\,°C at 22.8\,A are measured (direct current). Both currents far exceed the typical rating of \textit{RS550} DC motors, which typically lies around 5\,-\,8\,A. It is thought that the exceptional cooling provided by the propellers is largely responsible for the absence of overheating.

%%%%%%%%%%%%%%%%%%%%%%%%%%%
\vspace{14pt}
\section{Control Allocation}
When in flight, the position and orientation control of MOMAV is handled by a manually tuned PID controller. It takes as input the current position/orientation errors of the drone and outputs desired force/torque setpoints that act to reduce said errors. The way in which these setpoints are then allocated to propeller-throttle and arm-angle setpoints, however is novel. This algorithm for control allocation is presented below.
\vspace{12pt}\par 
\begin{center}
	\begin{tabular}{ l c l l }
		\hline
		& Name & Description & Type \\ 
		\hline
		\\[-8pt] \multicolumn{4}{l}{\textbf{Inputs:}} \\
		& $q$ & Body orientation quaternion & $\in \mathbb{H}$ \\
		& $F$ & Desired body force & $\in \mathbb{R}^3, [N]$ \\
		& $M$ & Desired body torque & $\in \mathbb{R}^3, [Nm]$ \\
		\\[-8pt] \multicolumn{4}{l}{\textbf{Outputs:}} \\
		& $a_i$ & Angle of arm $i$ & $\in \mathbb{R}, [rad]$ \\
		& $u_i$ & Motor throttle of arm $i$ & $\in \mathbb{R}, (0,1)$ \\
		\\[-8pt] \multicolumn{4}{l}{\textbf{Constants:}} \\
		& $r_i$ & Arm endpoint & $\in \mathbb{R}^3, [m]$ \\ 
		& $x_i$ & Arm rotation axis & $\in \mathbb{R}^3, \lVert \cdot \rVert=1$ \\
		& $z_i$ & Motor thrust direction at & $\in \mathbb{R}^3, \lVert \cdot \rVert=1$ \\
		& & zero arm angle & \\
		& $s_i$ & Motor spin direction & $\in \{-1, 1\}$ \\
		& $\mu$ & Motor thrust constant & $\in \mathbb{R}, [N/1]$ \\
		& $\tau$ & Motor torque constant & $\in \mathbb{R}, [Nm/1]$ \\
		& $\Delta t$ & Control loop period & $\in \mathbb{R}, [s]$ \\
		\\[-8pt] \multicolumn{4}{l}{\textbf{Intermediaries:}} \\
		& $n_i$ & Motor thrust direction & $\in \mathbb{R}^3, \lVert \cdot \rVert=1$ \\
		& $f_i$ & Motor force & $\in \mathbb{R}^3, [N]$ \\
		& $m_i$ & Motor torque & $\in \mathbb{R}^3, [Nm]$ \\
		\\[-8pt]
		\hline
	\end{tabular}
	\captionof{table}{Variable names and descriptions} 
\end{center}
\vspace{10pt}\par
MOMAV is highly over-actuated, featuring 12 actuators (6 for arm-angles and 6 for propeller-throttles), but only 6 constraints (3 for the force setpoint and 3 for the torque setpoint). It means that the constraints stemming from the desired force ($F$) and torque ($M$) are not sufficient to fully define the arm-angles ($a$) and propeller-throttles ($u$). For this reason, an optimization approach is used (Eq. \ref{eq:3_optim_prob}) where the force/torque constraints must be satisfied ($\mcomp G=0$) and the remaining variables are chosen such that they minimize some objective function ($O$).
\par
\begin{align}
	\begin{aligned}
		(u^*,a^*) = \ \mathrm{\mathbf{argmin}} \ &O(u,a) \\
		\mathrm{\mathbf{subject\ to}} \ &G(u,a,q,F,M)=0
	\end{aligned}
	\label{eq:3_optim_prob}
\end{align}
\vspace{0pt}\par
\vspace{6pt}\par
\begin{figure}[h!]
	\includegraphics[width=\columnwidth]{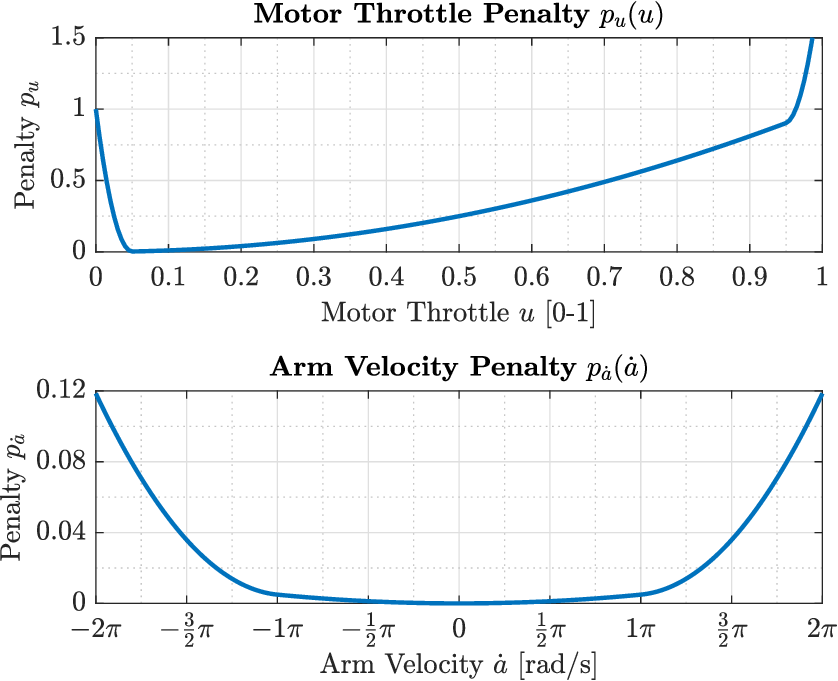}
	\centering
	\caption{Penalty functions used in optimization objective $O$}
	\label{fig:3_sqp_penalty}
\end{figure}

The objective function $O$ is chosen with the intention of minimizing power consumption, while keeping the solution ($u^*$, $a^*$) feasible. It is split up in four goals, each one weighted against the others:
\begin{itemize}
	\vspace{4pt}
	\item Minimize the squared sum of throttles
	\item Minimize the squared sum of arm rotation velocities
	\item Constrain the throttles within 0\,-\,100\%
	\item Constrain the arm rotation velocities within $\pm$1\,rev/sec
	\vspace{8pt}
\end{itemize}
The last two goals are inequality constraints, but since those can be cumbersome to deal with, they are instead implemented as quadratic penalty terms in the objective $O$. The resulting objective function $O$ (Eq. \ref{eq:O}) thus becomes a continuous piece-wise polynomial function, summing over penalty functions for throttles $p_u(u)$ and arm velocities $p_{\dot{a}}(\dot{a})$ (Fig. \ref{fig:3_sqp_penalty}).
\par
\begin{equation}
	O = \sum_{i=1}^{6} \ p_u(u_i) + \sum_{i=1}^{6} \ p_{\dot{a}}\left(\frac{a_i-a_{i,\mathrm{prev}}}{\Delta t}\right)
	\label{eq:O}
\end{equation}
\vspace{0pt}\par
\par
The constraint function $G$ is defined as the difference between the desired body force/torque $F, M$ and the sum of forces/torques $f_i,m_i$ achieved by the inputs $u_i,a_i$ (Eq. \ref{eq:G}). When these are equal to each other, it follows that $\mcomp G=0$ and the constraint is satisfied. Note that in the following equations quaternions are multiplied with vectors and the result is defined to be the rotated vector.
\par
\begin{align}
	n_i &= \m{ \sin(a_i/2) \\ x_i \cos(a_i/2) }_{\in \mathbb{H}} z_i \\
	f_i &= \mu \, u_i n_i \\
	m_i &= \mu \, u_i (r_i \times n_i) + \tau \, s_i u_i n_i \\[10pt]
	G &= \m{ \sum_{i=1}^{6} (f_i) - q^{-1}F \\ \sum_{i=1}^{6} (m_i) - q^{-1}M }
	\label{eq:G}
\end{align}
\vspace{0pt}\par
Now that the optimization problem is fully defined, a few characteristics can be noted:
\par
\begin{itemize}
	\item The problem is nonlinear due to the trigonometric functions in the constraint $G$, suggesting that a nonlinear iterative solver is required.
	\item The objective $O$ is a convex function, meaning there is no risk of converging to a local maximum instead of a local minimum.
	\item The constraint $G$ is not convex, meaning a solver could converge to a sub-optimal solution that for example asks an arm to rotate by 360°. Although in practice this can be easily avoided.
\end{itemize}
\par
The numerical method chosen to find the optimal solution $u^*$ and $a^*$ is sequential quadratic programming (SQP) \cite{1999_wright_numop}. In an effort to make the resulting equations more understandable, they are  derived by applying Newton's method to the first-order optimality conditions, which are:
\par
\begin{itemize}
	\item \textbf{Primal Feasibility:} $\mcomp G = 0$, stating that the constraint must be satisfied at the solution.
	\item \textbf{Stationarity:} $\mcomp \nabla O + \nabla G^T \lambda = 0$, stating that the gradient of the objective must be a linear combination of the gradients of every constraint (linked by some factors $\lambda$). This ensures that no direction exists, that is orthogonal to every constraint gradient, and at the same time is not orthogonal to the objective gradient. Moving in such a direction would otherwise allow the objective to be lowered while maintaining constraint satisfaction.
\end{itemize}
\par
These two equations are stacked into one and the derivatives split into parts with respect to $u$ and $a$ (Eq. \ref{eq:dL}). Written in this form, the equation is equivalent to stating that the gradient of the Lagrangian $\mcomp\mathcal{L}=O+G^T\lambda$ must be zero: $\mcomp\nabla\mathcal{L}=0$. Below, all derivatives are written as subscripts, e.g. $\mcomp O_u = \nabla_u O = \partial O/\partial u$. Their definitions are found in appendix \ref{sec:deriv_SQP}.
\par
\begin{align}
	\nabla\mathcal{L}(u^*,a^*,\lambda^*) = \m{O_u + {G_u}^T \lambda \\ O_a + {G_a}^T \lambda \\ G} _{\substack{u=u^*\\a=a^*\\\lambda=\lambda^*}}= 0
	\label{eq:dL}
\end{align}
\vspace{0pt}\par
To arrive at the SQP algorithm, the above equation is solved iteratively with Newton's method, by taking a linear approximation at the current best guess $[u,a,\lambda]$ (Eq. \ref{eq:dL_approx}), and then finding an improvement step $[\delta u, \delta a, \delta \lambda]$ which solves that approximated equation (Eq. \ref{eq:step}).
\par
\begin{align}
	&\nabla\mathcal{L}(u+\delta u,a+\delta a,\lambda+\delta \lambda) \approx H\m{\delta u \\ \delta a \\ \delta \lambda} + K = 0 \label{eq:dL_approx}\\
	&\qquad H = \m{O_{uu} + G_{uu} : \lambda & O_{ua} + G_{ua} : \lambda  & {G_u}^T \\ O_{au} + G_{au} : \lambda & O_{aa} + G_{aa} : \lambda  & {G_a}^T \\ G_u & G_a & 0} \label{eq:H}\\
	&\qquad\qquad\ \textrm{with:}\ (G_{\star \bullet}:\lambda)_{ij} = (\partial^2 G / \partial \star_i \partial \, \bullet_j )^T\lambda\ ) \nonumber\\
	&\qquad K = \m{O_u + {G_u}^T \lambda \\ O_a + {G_a}^T \lambda \\ G} \label{eq:K}\\
	&\m{\delta u \\ \delta a \\ \delta \lambda} = H^{-1}K \label{eq:step}
\end{align}
\vspace{0pt}\par
Iteratively updating the current best guess by the improvement step $\mcomp [u,a,\lambda] \leftarrow [u,a,\lambda] + [\delta u, \delta a, \delta \lambda]$ should then converge to a solution $[u^*,a^*,\lambda^*]$, where $u^*$, $a^*$ are such that they produce the desired force/torque setpoints $F$,$M$, and where the leftover degrees of freedom are such that they minimize the objective function $O$.
\par
Although, usually in SQP only a scaled down improvement step $\mcomp\alpha\cdot[\delta u, \delta a, \delta \lambda]$ is taken, with $\mcomp \alpha\in[0,1]$ determined by a line-search algorithm and some merit function. For this particular problem it was found to be sufficient to pick $\alpha$ such that no single arm-angle nor throttle step exceeds a fixed limit (Eq. \ref{eq:alpha}).
\par
\begin{align}
	\alpha = \mathrm{min}(1.0, \frac{\delta a_{lim}}{\mathrm{max}(\delta a)}, \frac{\delta u_{lim}}{\mathrm{max}(\delta u)}) \label{eq:alpha}
\end{align}
\vspace{0pt}\par
On MOMAV, the complete SQP based control allocation algorithm (Alg. \ref{alg:sqp}) is implemented as a ROS node in C++ and executed on a \textit{Radxa Rock 5A} computer, serving as the flight controller. The \textit{PartialPivLU} solver from the \textit{Eigen} C++ library is used to solve the linear system of equations (Eq. \ref{eq:step}). This setup manages to converge to a solution reliably within 1\,-\,8 iterations, corresponding to 0.4\,-\,3.2\,ms in computation time. It manages to do so even during fast orientation changes, when the solution is furthest from the initial guess.
\vspace{6pt}\par
\begin{algorithm}
	\caption{SQP algorithm}\label{alg:sqp}
	\begin{algorithmic}
		\Require{$q, F, M$}
		\Ensure{$u, a, \lambda$}
		\State $u,a,\lambda \gets \text{result from previous run}$
		\Repeat
		\State $H,\;K \gets \text{calculate with (\ref{eq:H}), (\ref{eq:K})}$
		\State $(\delta u, \delta a, \delta \lambda) \gets \text{solve}\ H^{-1}K$
		\State $\alpha \gets \text{calculate with (\ref{eq:alpha})}$
		\State $(u,a,\lambda) \gets (u,a,\lambda) + \alpha (\delta u, \delta a, \delta \lambda)$
		\State $O,\;G \gets \text{calculate with (\ref{eq:O}), (\ref{eq:G})}$
		\Until{$\lvert O-O_{\mathrm{prev}} \lvert\:/\:O < \text{tol.} \ \textbf{and} \  \lVert G \rVert < \text{tol.} $}
		\State \Return $u,a,\lambda$
	\end{algorithmic}
\end{algorithm}
\par
\begin{figure*}[b!]
	\vspace{12pt}
	\includegraphics[width=\textwidth]{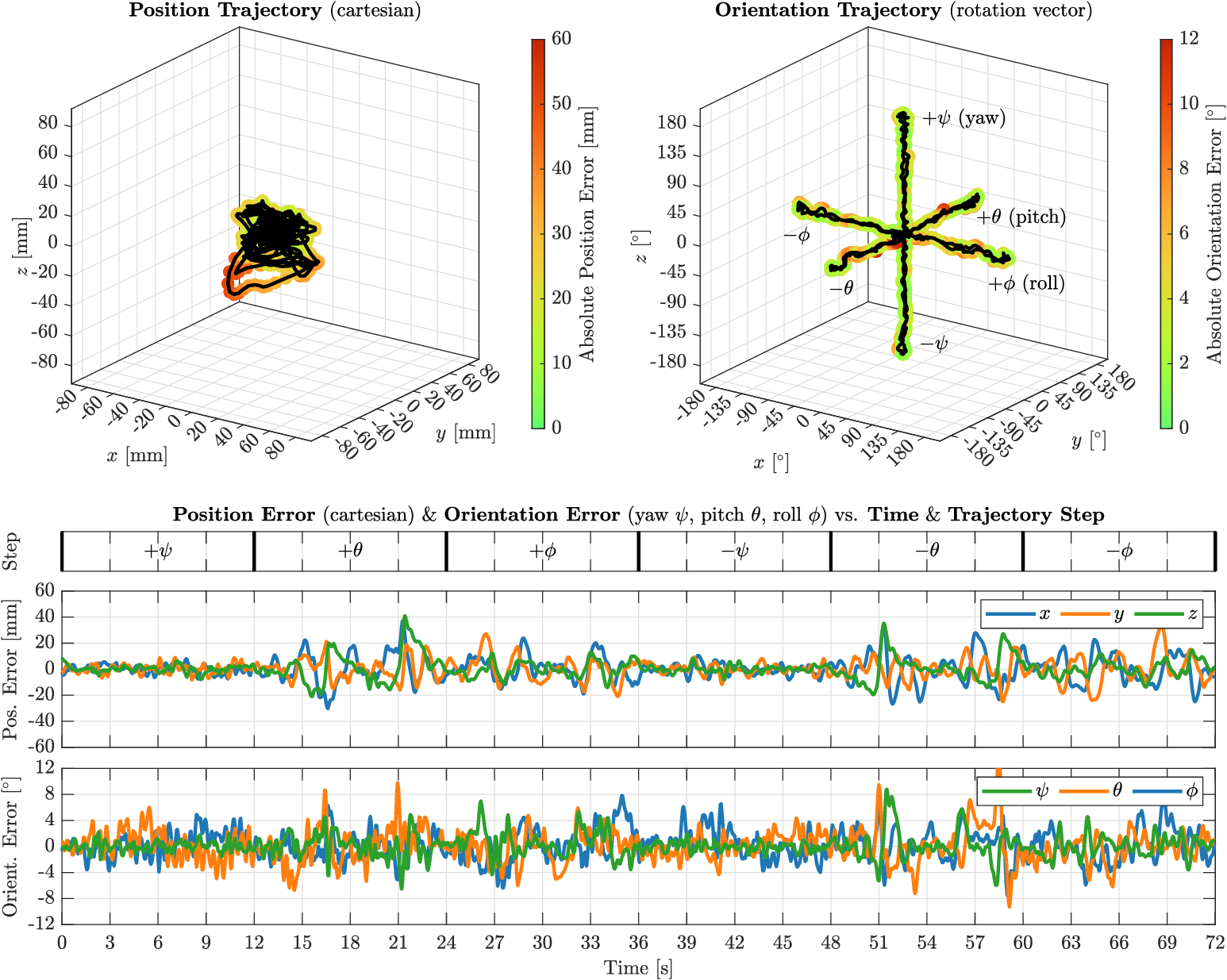}
	\centering
	\caption{Test flight sweeping through orientations and keeping the position fixed}
	\label{fig:4_flight_rot_sweep}
	\vspace{-4pt}
\end{figure*}
\par
The main disadvantages of this algorithm is that it does not come with any guarantees on computation time, due to its iterative nature. Although this problem can be successfully mitigated by a fast computer, there also exist other algorithms based on the Moore-Penrose pseudoinverse method, implemented in \cite{2020_xu_bicopter, 2019_bodie_omav, 2018_kamel_voliro, 2015_ryll_quadvc, 2013_sequigasco_quadvcd}, which do not suffer from the same issue.
\par
Its main advantage on the other hand is that the behavior of the control allocation can be easily adjusted by modifying the terms in the objective function $O$. For example, a common challenge in variable-tilt drones is a singularity in arm-angle when an arm happens to point along the vertical axis during hover. The pseudoinverse based control allocation could demand that arm to instantly rotate 180° in such a situation, which is unfeasible. By penalizing arm-angle velocities in the objective $O$ this issue is easily avoided in the here presented SQP based algorithm. Also, since the SQP method works with nonlinear problems, it could potentially be extended to account for other challenging nonlinear effects such as angle dependent slip-ring losses, propeller wake interactions, propeller slew rates, and more.

%%%%%%%%%%%%%%%%%%%%%%%%%%%
\begin{figure*}[t!]
	\includegraphics[width=\textwidth]{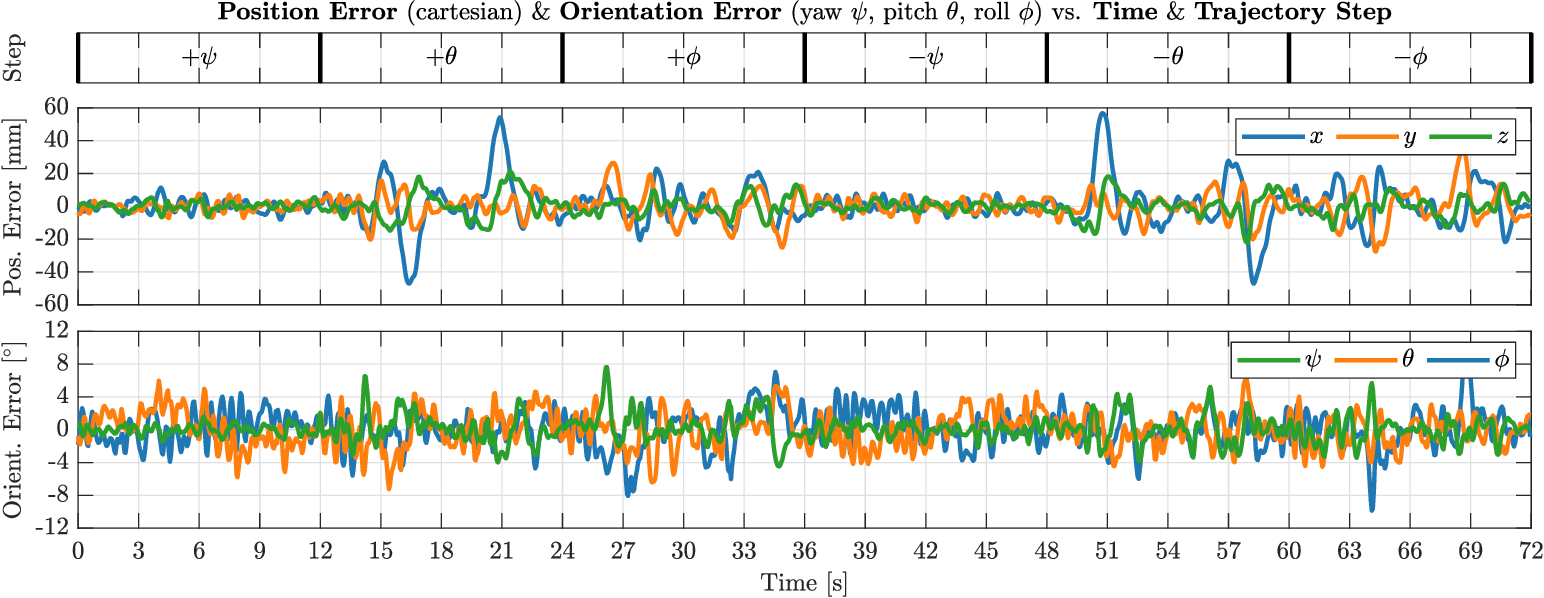}
	\centering
	\caption{Test flight with a trajectory like in Fig. \ref{fig:4_flight_rot_sweep}, but using the Moore-Penrose pseudoinverse control allocation method}
	\label{fig:4_flight_no_sqp}
	\vspace{-4pt}
\end{figure*}
\begin{figure*}
	\vspace{16pt}
	\includegraphics[width=\textwidth]{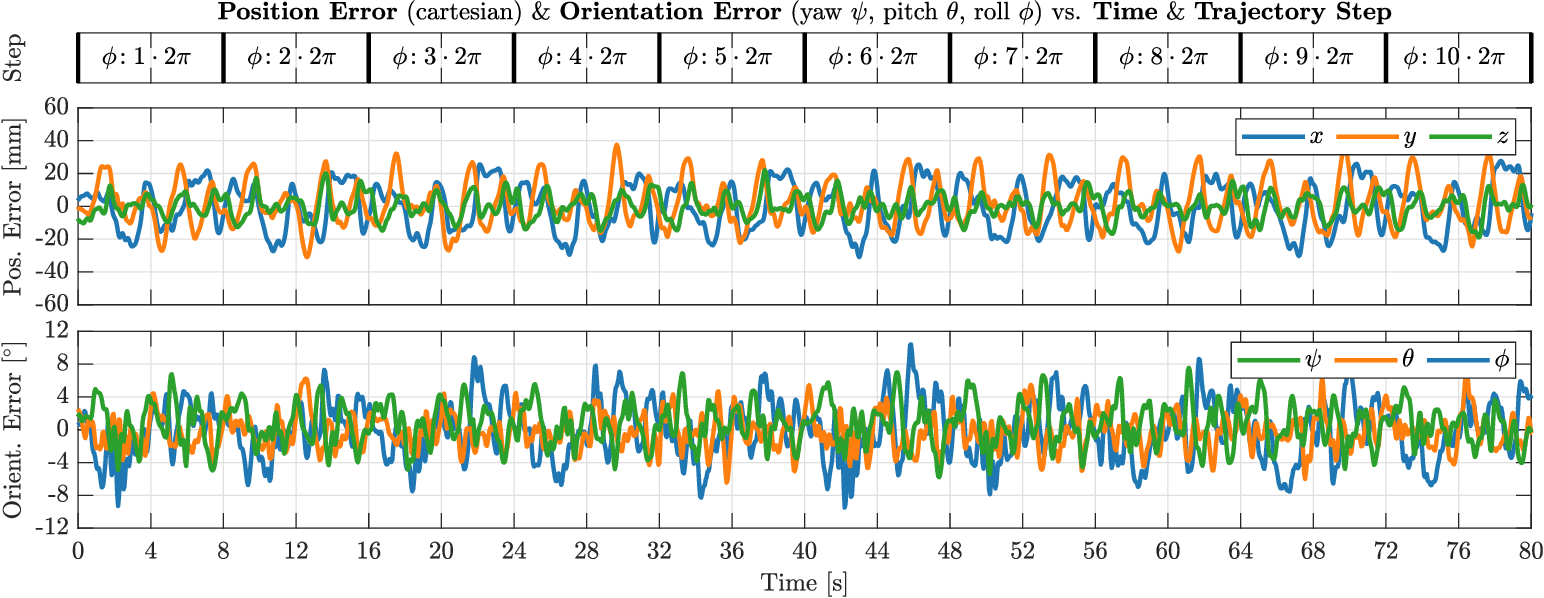}
	\centering
	\caption{Test flight with ten revolutions in roll while maintaining the position stationary}
	\label{fig:4_flight_cont_rot}
	\vspace{-4pt}
\end{figure*}
\begin{figure*}
	\includegraphics[width=\textwidth]{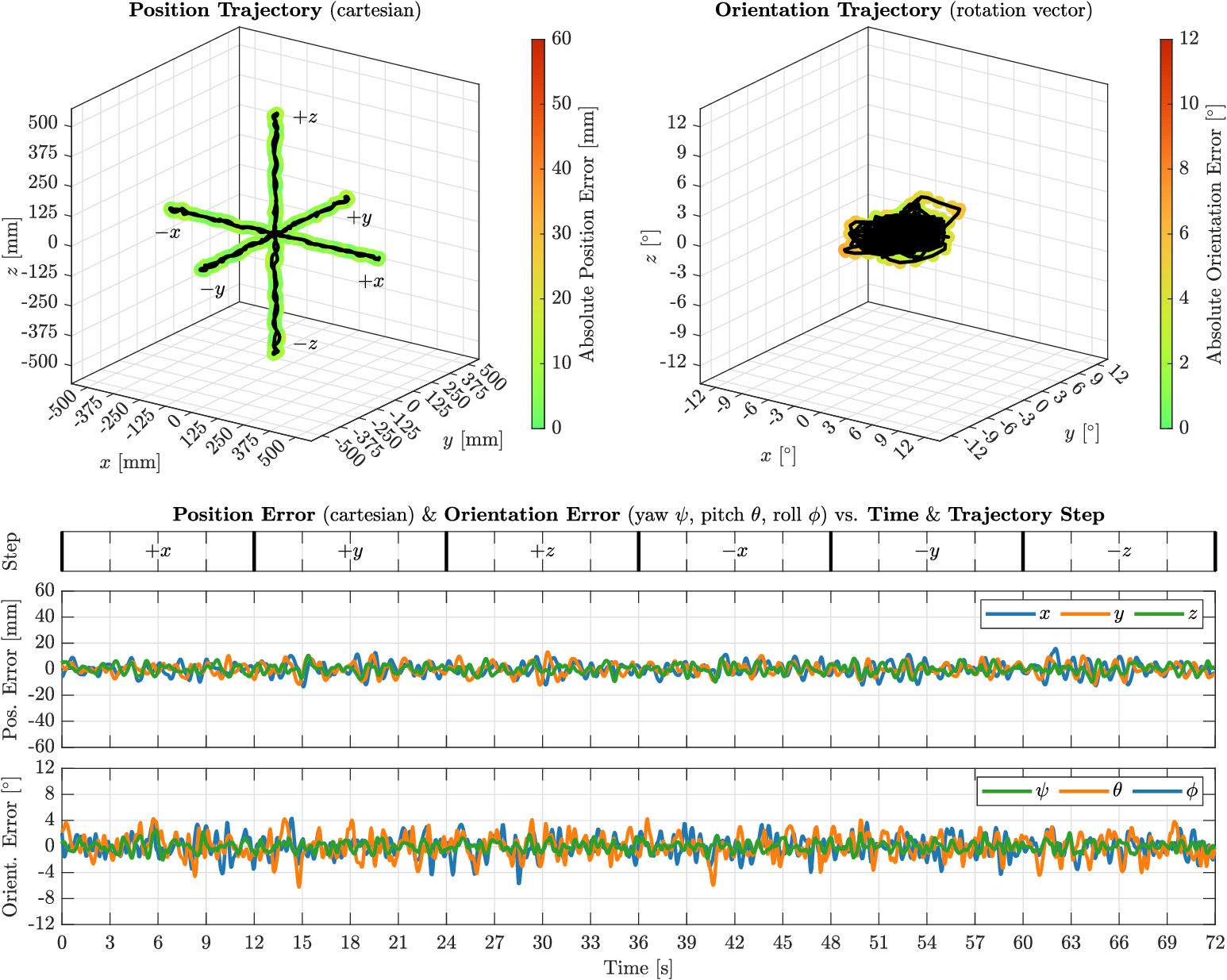}
	\centering
	\caption{Test flight sweeping through positions and keeping the orientation fixed}
	\label{fig:4_flight_pos_sweep}
	\vspace{-4pt}
\end{figure*}
\vspace{8pt}
\section{Test Flights}
Finally, test flights are presented, to show how well MOMAV performs in reality. An \textit{OptiTrack Prime\textsuperscript{x} 13W} tracking system is used to provide ground-truth position and orientation data for all tests. The same tracking data is also used on-board for control feedback during flight.
\par
Two types of test flights are performed to show MOMAV's ability to decouple position from orientation. In one, the orientation setpoint is held fixed, while the position setpoints sweep along the $x$, $y$, $z$ axes. In the other, the position setpoint is held fixed, while the orientation setpoints sweep along the yaw ($\psi$), pitch ($\theta$), roll ($\phi$) axes. The sweeps are sequenced with each axis, in positive/negative direction, and always returning to the original pose in between. For example, the orientation-sweep steps are: $0, +\psi, 0, +\theta, 0, +\phi, 0, -\psi, 0, -\theta, 0, -\phi$. Orientation steps reach up to $\pm$180° and position steps reach up to $\pm$500\,mm, taking 6\,s per step with smooth accelerations/decelerations equal to half the maximum velocity.
\par
During the orientation-sweep test flight (Fig. \ref{fig:4_flight_rot_sweep}) a mean orientation error of 3.3° is measured ($\sigma$: 2.0°, P\textsubscript{90}: 5.9°), as well as a mean position error of 11.8\,mm ($\sigma$: 8.6\,mm, P\textsubscript{90}: 24.2\,mm). The orientation error appears to correlate with angular velocity in pitch and roll, which is maximal around seconds 15, 21, 27, 33, 51, 57, 63, 69. Since at other times the integrating term of the PID orientation controller has more time to offset static errors, it is thought that the error stem from inaccuracies in the thrust/torque model, possibly due to wake interactions or due to arm or angle dependent slip-ring losses.
\par
A comparison with an orientation-sweep test flight that uses the Moore-Penrose pseudoinverse control allocation instead of the novel SQP based method (Fig. \ref{fig:4_flight_no_sqp}) shows peaks in position errors at seconds 15, 21, 51, 57. These are also the times where some arms align with the vertical axis and arm-angle singularities occur. Statistical analysis of 12 such instances shows a decrease in mean peak position error from 52.6mm to 43.2mm, with a significance of $p=0.0003$. This decrease is an advantage of using the SQP based control allocation method with arm-angle velocity penalty.
\par
Furthermore, a slightly modified orientation-sweep test flight is presented to showcase the usefulness of the slip-rings (Fig. \ref{fig:4_flight_cont_rot}). In this flight the roll axis is swept for 10 revolutions at 8 seconds per revolution without interruptions. The slip-rings enable such a flight by preventing motor cables from winding up around the arms. Analysis of the flight shows a very periodic behavior of the position error, while the orientation error appears to be more random. The apparent correlation between roll angle and position error again indicates a possible oversimplification in the motor thrust/torque model.
\par
Finally, a position-sweep test flight is also presented \mbox{(Fig. \ref{fig:4_flight_pos_sweep})}. Compared to the orientation-sweep flight it shows a lower mean orientation error of 2.1° ($\sigma$: 1.0°, P\textsubscript{90}: 3.6°), as well as a much lower mean position error of 6.6\,mm ($\sigma$: 3.0\,mm, P\textsubscript{90}: 10.6\,mm). No correlation between setpoints and errors is apparent in this test.
\vspace*{4pt}

%%%%%%%%%%%%%%%%%%%%%%%%%%%
\section{Conclusion}
MOMAV was able to showcase multiple interesting findings. It showed that an octahedral arm configuration with rotating arms is able to hover in any orientation, while wasting little thrust to motors having to fight each other, and requiring little surplus total thrust in the drone design compared to other fully-actuated drones. The rotating arms showed the viability of using modified hobbyist servos as performant actuators, supporting precise angular control across revolutions. They also showed the viability of using DC motor brushes as high current slip-rings, and their potential to incur only negligible power losses. A SQP based algorithm was furthermore introduced as a highly tunable method for control allocation, specifically tailored to drones with rotating arms.
\par
All these findings were put to the test during actual flights of the prototype drone. They showed MOMAV's ability to control its orientation independently of its position and vice-versa.  They also showed a merit of the SQP based control allocation in suppressing arm angle singularities, and they showed MOMAV's ability to rotate indefinitely thanks to the slip-rings on its arms.
\par
For readers interested in further information: An overview of the control architecture, as well as a detailed descriptions of each of its parts is available under \textit{\url{\urlinfo}}. The entire project source-code, 3D models of the drone, and raw data of all tests are also freely available under \textit{\url{\urlgit}}.
\par
Future development of MOMAV is not planned. If it were, it would include the implementation of a nonlinear motor thrust model, or even a learned model based on in-flight system identification. It is thought that this change could further reduce position and orientation tracking error by accounting for effects like wake interactions. Also of interest would be tests showing how well the tracking accuracy in free flight translates to accuracy in manipulation tasks involving contact with the environment.
\par
This research did not receive any specific grant from funding agencies in the public, commercial, or not-for-profit sectors.

%%%%%%%%%%%%%%%%%%%%%%%%%%%
\vspace{8pt}
\appendix
\vspace{6pt}
\subsection{Drone Characteristics}
\begin{tabular}{ l l }
	&\\[-8pt]  \hline  &\\[-8pt]
	Weight with batteries & 2.4 kg \\
	Weight without batteries & 1.6 kg \\
	Upwards Thrust (80\% throttle) & \txtapprox 4.6 kg \\
	Diagonal with propellers & 590 mm \\
	Diagonal without propellers & 412 mm \\
	Hover flight time & 11 min\\
	&\\[-8pt]  \hline  &\\[-7pt]
	Maiden flight & 9 May 2022\\
	Published flights & 6 April 2024\\
	Test flight count & 341\\
	Crash count & 28\\
	Development time & 8 months\\
	Components cost & 2'200€\\
	&\\[-8pt]  \hline &\\
\end{tabular}
\vspace{6pt}
\subsection{Drone Components}
\begin{tabular}{ l l l }
	\hline
	&&\\[-8pt] \multicolumn{3}{l}{\textbf{Core:}} \\
	& Computer & Radxa Rock 5A \\
	& Tracking & Intel RealSense T265 or\\
	&& OptiTrack Prime\textsuperscript{x} 13W \\
	& Power & I3A4W 008A033V (5V)\\
	&& I6A4W 020A033V (8.4V)\\
	& Battery & Swaytronic 6S 2800mAh (2x) \\
	&&\\[-8pt] \multicolumn{3}{l}{\textbf{Arm Assembly:}} \\
	& Servo & KST MS325 \\
	& Controller & Atmel SAM D21 \\
	& Driver & TI DRV8870 \\
	& Brushes & RS550 DC Motor Brushes \\
	&&\\[-8pt] \multicolumn{3}{l}{\textbf{Propulsion:}} \\
	& Motor & T-Motor F100 \\
	& Driver & T-Motor MINI F45A 6S 4In1 (2x)\\
	& Propeller & DJI Mavic Pro 8331 (3-blade-mod)\\
	&&\\[-8pt] \multicolumn{3}{l}{\textbf{Structural:}} \\
	& Sheets & Swiss-Composite 1.5mm CFRP Sheet \\
	& Arm Tubes & Swiss-Composite 8x10mm CFRP Tube \\
	& 3D Parts & Dutch Filaments PET-G with Carbon \\
	&&\\[-8pt] \hline &&\\
\end{tabular}
%\vspace{-12pt}
\subsection{Derivatives used in control allocation}
\label{sec:deriv_SQP}
\vspace{-14pt}
\allowdisplaybreaks
\begin{alignat*}{3}
	\hline & \mspace{160mu} && \mspace{160mu} \\[-8pt]
	&\pmb{\frac{\partial n_i}{\partial a_i}} = x_i \times n_i
	&&\mathrlap{\pmb{\frac{\partial^2 n_i}{\partial {a_i}^2}} = x_i \times (x_i \times n_i)} \\[12pt]
	&\pmb{\frac{\partial f_i}{\partial u_i}} = \mu n_i
	&&\pmb{\frac{\partial^2 f_i}{\partial {u_i}^2}} = 0
	&&\pmb{\frac{\partial^2 f_i}{\partial u_i \partial a_i}} = \mu \frac{\partial n_i}{\partial a_i} \\[4pt]
	&\pmb{\frac{\partial f_i}{\partial a_i}} = \mu u_i \frac{\partial n_i}{\partial a_i}
	&&\pmb{\frac{\partial^2 f_i}{\partial {a_i}^2}} = \mu u_i \frac{\partial^2 n_i}{\partial {a_i}^2} \\[12pt]
	&\mathrlap{\pmb{\frac{\partial m_i}{\partial u_i}} = \mu (r_i \times n_i) + \tau s_i n_i}
	&& &&\pmb{\frac{\partial^2 m_i}{\partial {u_i}^2}} = 0 \\[4pt]
	&\mathrlap{\pmb{\frac{\partial m_i}{\partial a_i}} = \mu u_i (r_i \times \frac{\partial n_i}{\partial a_i}) + \tau u_i s_i \frac{\partial n_i}{\partial a_i}} \\[4pt]
	&\mathrlap{\pmb{\frac{\partial^2 m_i}{\partial {a_i}^2}} = \mu u_i (r_i \times \frac{\partial^2 n_i}{\partial {a_i}^2}) + \tau u_i s_i \frac{\partial^2 n_i}{\partial {a_i}^2}} \\[4pt]
	&\mathrlap{\pmb{\frac{\partial^2 m_i}{\partial u_i \partial a_i}} = \mu (r_i \times \frac{\partial n_i}{\partial a_i}) + \tau s_i \frac{\partial n_i}{\partial a_i}} \\[8pt]
	&\pmb{\frac{\partial G}{\partial u_i}} = \m{\partial f_i / \partial u_i \\ \partial m_i / \partial u_i}
	&&\ \mathrlap{\pmb{\frac{\partial^2 G}{\partial u_i \partial u_j}} = \begin{cases} i=j: & \m{\partial^2 f_i / \partial {u_i}^2 \\ \partial^2 m_i / \partial {u_i}^2} \\ i\neq j: & 0 \end{cases}} \\[2pt]
	&\pmb{\frac{\partial G}{\partial a_i}} = \m{\partial f_i / \partial a_i \\ \partial m_i / \partial a_i}
	&&\ \mathrlap{\pmb{\frac{\partial^2 G}{\partial a_i \partial a_j}} = \begin{cases} i=j: & \m{\partial^2 f_i / \partial {a_i}^2 \\ \partial^2 m_i / \partial {a_i}^2} \\ i\neq j: & 0 \end{cases}} \\[2pt]
	&\mathrlap{\pmb{\frac{\partial^2 G}{\partial u_i \partial a_j}} = \pmb{\frac{\partial^2 G}{\partial a_j \partial u_i}} = \begin{cases} i=j: & \m{\partial^2 f_i / \partial u_i \partial a_i \\ \partial^2 m_i / \partial u_i \partial a_i} \\ i\neq j: & 0 \end{cases}}%\\[8pt]
\end{alignat*}
\vspace{12pt}
\bibliography{paper}
	
\end{document}